\documentclass[11pt]{article}
\usepackage[margin=1in]{geometry}
\usepackage{titlesec}
\titleformat*{\section}{\large\bfseries}
\titleformat*{\subsection}{\bfseries}
\titleformat*{\subsubsection}{\itshape}
\usepackage{comment}
\usepackage[title]{appendix}
\usepackage{booktabs} 
\usepackage{tabularx}
\usepackage{float} 
\usepackage{makecell}
\usepackage[authoryear,round]{natbib}
\usepackage{authblk}

\usepackage{amsmath, amssymb, siunitx}
\usepackage{bm, bbm}
\DeclareMathAlphabet{\mathbbm}{U}{bbold}{m}{n}

\usepackage{graphicx}
\usepackage{booktabs}
\usepackage[font=small, skip=4pt]{caption}
\usepackage[skip=4pt]{subcaption}
\usepackage[table]{xcolor}

\usepackage{enumitem}
\setlist[itemize]{nosep}

\usepackage[colorlinks, allcolors=blue, pdfstartview=]{hyperref}
\usepackage[nameinlink, capitalize]{cleveref}
\usepackage{natbib}
\title{Generative AI for urban planning: synthesizing land-use maps with satellite imagery via DualControlNet} 
\author{
Qingqi Song\textsuperscript{1},
Yuzhou Chen\textsuperscript{1},
Lingqian Hu\textsuperscript{2},
Shenhao Wang\textsuperscript{1,*}
}

\begin{document}
\maketitle

\vspace*{-2em}
\begin{center}
\textsuperscript{1} Department of Urban and Regional Planning, University of Florida\\
\textsuperscript{2} College of Architecture, Texas A\&M University\\
\end{center}
\date{}

\maketitle
\begin{abstract}
\noindent Various land-use models have been developed to characterize land-use patterns and facilitate land-use scenario planning. However, existing land-use models are largely predictive, forecasting land-use transitions based on numeric data, lacking the capability to synthesize non-existing land-use maps from simple text prompts. Meanwhile, generative AI is increasingly applied to urban systems, but existing generative approaches remain unimodal, weakly controllable, and evaluated mostly by generic image quality rather than domain-specific land-use metrics. To address these gaps, we introduce DualControlNet, a generative land-use model that jointly synthesizes high-resolution land-use maps and satellite imagery, simultaneously addressing the challenges in multimodality, controllability, and verification. Empirically, we train DualControlNet on more than 40,000 paired tiles from Orlando and Chicago and evaluate it using land-use, spatial consistency, and image quality metrics. Relative to unimodal ControlNet baselines, DualControlNet improves the consistency in land-use diversity by 62.8\% and in building coverage ratio by 16.3\%, more than doubles cross-modal spatial consistency, and improves visual quality by around three times. Through scenario-based generative experiments, we demonstrate that DualControlNet can support sustainable land-use development through multidimensional controls. It creates land-use maps that successfully increase land-use density and diversity, and respond to textual land-use prompts, such as ``high-density development'', thus assisting land-use scenario exploration through intuitive textual and imagery formats. Overall, this work advances land-use modeling by introducing a generative framework that is multimodal, controllable, and verifiable using land-use metrics. It further demonstrates the potential of generative AI for lowering the technical barrier to land-use scenario planning and enabling more intuitive and participatory exploration of urban futures.

\smallskip
\noindent *Corresponding author: Shenhao Wang; shenhaowang@ufl.edu
\end{abstract}

\noindent \textbf{Acknowledgments}: The corresponding author acknowledges the support from Research Opportunity Seed Grant (ROSF-2023) and Artificial Intelligence and Informatics Research Institute Seed Grant (AIIRI-2026) at the University of Florida.

\newpage
\section{Introduction}
Land-use patterns crucially shape urban sustainability, health, equity, and safety \citep{cervero_travel_1997, ewing_travel_2010, lyu_investigating_2025, weng_15-minute_2019, gu_density_2022, kim_density_2021}. To characterize land-use patterns and inform planning practices, researchers have developed a rich set of land-use metrics, including land-use diversity, density, and building coverage, together with land-use models that forecast how urban land use evolves over time \citep{boeing_osmnx_2017, He2023, Lopane2023}. These land-use models, ranging from cellular automata to statistical methods, enable researchers to predict land-use transitions using numeric and historical land-use data \citep{CruzBello2024, Guan2024, Cai2026}. However, they share a common limitation. As they are largely predictive models, they forecast incremental transitions from existing land-use categories, but cannot synthesize land-use patterns that do not yet exist. This limitation matters for scenario planning, a core planning practice in which planners envision and compare alternative urban futures before committing to a specific plan. Such practice requires the capacity to generate new land-use maps directly from planning objectives, rather than only extrapolating from the past. 

Recently, generative AI introduces an opportunity of directly synthesizing high-fidelity spatial representations from abstract inputs such as textual prompts \citep{ho2020denoising, rombach_high-resolution_2022}. However, the planning research community still lacks a verifiable, controllable, and multimodal generative framework for land-use scenario planning. First, most generative studies remain unimodal, synthesizing a single urban layer such as satellite imagery or a building footprint map in isolation \citep{wu2022ganmapper,  dedring2024synthesis, wang2025generative, sun_envisioning_2026, he_human-guided_2026}. This is a fundamental limitation because urban systems are inherently multimodal\footnote{The term ``multimodal'' refers to multiple urban system layers in this study, different from the multimodal data structures commonly used in the AI community.}: land use, urban form, road networks, and other layers align spatially and should be generated jointly to preserve their cross-modal consistency. Second, existing models offer weak domain-specific controllability, providing limited mechanisms to steer generation using interpretable land-use metrics such as land-use diversity or building density \citep{lowry2014comparing, abou2022overview}. Yet such control is essential for scenario planning, where planners specify explicit development objectives and expect matching outputs. Third, generated outputs are typically evaluated only by generic image realism metrics, which reveal whether an image looks plausible but not whether it is geographically valid against targeted land-use metrics \citep{castro2025comprehensive, ye2022masterplangan, abdollahi2020building}. Therefore, it is crucial to envision a generative land-use model that synthesizes land use jointly with other urban modalities, responds to interpretable planning objectives, and is evaluated against domain-specific land-use metrics. 

To address these gaps, we introduce DualControlNet, a generative land-use model that jointly synthesizes land-use maps and satellite imagery, controls generation with interpretable land-use metrics, and verifies the generated outputs against the same metrics. Technically, we assemble more than 40,000 observations from Orlando and Chicago, each aligning a satellite image, a land-use map, and a control map of major roads and water bodies. We compute land-use metrics such as building coverage ratio, land-use diversity, and average land-use area from every observation, and translate them into structured language prompts. We then extend the ControlNet architecture to a dual-modality setting and add a geospatial consistency loss that regularizes the boundary geometry shared by the two generated modalities. Such a research design allows us to address three research questions: first, whether the generated land-use maps satisfy the land-use metrics specified in the prompts; second, whether the generated land-use map and satellite imagery remain spatially consistent with each other; lastly, whether the generated satellite imagery is visually realistic for credible planning communication.

After the technical evaluation, we further use DualControlNet for generative scenario planning of sustainable land development. We start from uni-dimensional control, increasing the target building coverage ratio and examining whether the generated land-use map achieves the intended development intensity. We then move to multidimensional control, jointly raising building coverage and the number of land-use categories at underdeveloped sites. Such generative scenario planning is highly relevant to the cycle in planning practice: a planner states development objectives in plain language, receives a spatially coherent and visually inspectable land-use map within seconds, and then verifies that scenario against the very initial objectives.

This study is organized as follows. Section~\ref{sec:lit} reviews land-use metrics and modeling, generative AI for urban systems, and scenario planning, and identifies the research gaps. Section~\ref{sec:method} describes the multimodal data and introduces DualControlNet with its geospatial consistency loss. Section~\ref{sec:experiment} presents the experimental design and the three families of evaluation metrics. Section~\ref{sec:results} reports the model benchmarks and the generative scenario planning experiments. Section~\ref{sec:dis} discusses the implications for planning practice and the limitations. Section~\ref{sec:con} concludes our study.

\section{Literature Review}
\label{sec:lit}
\subsection{Land-use metrics and modeling}
Researchers have designed a wide range of land-use metrics to characterize land-use patterns and their implications for urban sustainability, health, equity, and safety. The land-use metrics include land-use diversity, which is typically measured by entropy-based indices and captures the functional heterogeneity of an area. They also include density, building coverage ratio, or average building size, which capture the intensity and compactness of development. The classic density–diversity–design (3D) framework established these as core dimensions of land use \citep{cervero_travel_1997}, and subsequent work extended them to more general urban form metrics, including street design, accessibility, transit proximity, and spatial-network properties \citep{ewing_travel_2010, boeing_osmnx_2017}. These land-use metrics matter because they can explain broad urban outcomes, such as sustainability, health, equity, and safety. For sustainability, the land-use metrics explain walking, transit use, and urban vitality \citep{aditjandra_exploring_2016, farahani_review_2013, lyu_investigating_2025}. For equity, they can measure spatial and social equity of people's accessibility especially around the elderly communities \citep{gu_density_2022}. For safety, the land-use metrics can measure the crime risk across street segments \citep{kim_density_2021}. Theoretical frameworks such as Lynch's legibility \citep{lynch_image_2008} and comprehensive urban design \citep{carmona_public_2021} further situate land-use metrics as diagnostic tools for inclusive, sustainable, and resilient urban development. Such land-use metrics have been widely incorporated into land-use modeling through regression approaches \citep{boarnet_broader_2011}, multi-source big data approaches \citep{zhang_systematic_2019, lyu_investigating_2025}, or recently machine learning methods \citep{li_effects_2024}. 

Diverse land-use models have been developed to understand and forecast urban land-use changes based on historical land-use patterns or sociodemographic drivers. The first land-use model family is cellular automata (CA) and its variants, which simulate land-use change through local transition rules, including partitioned and vector-based formulations that account for spatial heterogeneity and parcel structure \citep{He2023, CruzBello2024, YangVCA2023, Yao2024}. Recent studies also expanded the classical CA approach by integrating deep learning architectures such as convolutional neural networks (CNNs) and graph neural networks (GNNs) to capture spatial dependencies and nonlinear transition dynamics \citep{Guan2024, Xu2024}. Besides CA models, a second family relies on statistical or machine learning methods, ranging from dynamic spatial regression \citep{Tepe2023} to deep learning architectures that learn nonlinear transition patterns in an end-to-end manner \citep{WangUNet2022, Rath2026, WangXAI2025}. Along this line, researchers further developed models that integrate transportation systems \citep{Lopane2023}, cooperative models \citep{Zeng2025}, and vision-language models \citep{Cai2026}. Despite their methodological diversity, these models share a common logic: they are predictive models forecasting the transition of land-use categories based on historical data, mostly relying on incremental changes in land-use categories rather than directly synthesizing non-existing land-use patterns. 

\subsection{Generative AI for urban systems}
Generative AI is increasingly applied to geospatial analytics and urban planning tasks, enabling the synthesis of street layouts, building footprints, and urban imagery. Early work often leveraged GAN-based pipelines to support design automation and rapid visualization: MasterplanGAN and Urban-GAN generate masterplan-like urban layouts to streamline large-scale design workflows \citep{ye_masterplangan_2022, quan_urban-gan_2022}, and other approaches reconstruct missing or sparse geospatial datasets to alleviate data scarcity in mapping and planning support \citep{wu_generative_2022}. As summarized by \citet{jiang_generative_2024}, existing work highlights the promise of generative urban design in enhancing creativity, participation, and communication in the planning process. 

However, the generation of land-use maps imposes stronger requirements than generic image synthesis. Specifically, the research community still lacks a generative framework that can jointly generate multiple urban layers while remaining controllable and verifiable. First of all, urban systems are inherently multimodal: land use, road networks, building footprints, and other spatial representations are spatially aligned layers and must remain consistent when one spatial map is generated. Most of the existing generative literature, however, remains centered on producing a single layer and only loosely connecting to other modalities in downstream steps \citep{ye_masterplangan_2022, quan_urban-gan_2022, wu_generative_2022}. Emerging multimodal generative directions suggest feasibility but also underline the gap. For instance, multimodal approaches that combine visual and auditory perceptions extend what urban representations can include, offering richer depictions of human experience in cities \citep{zhuang_hearing_2024}. Meanwhile, unified or cooperative multimodal diffusion models for text--image and audio--visual generation improve cross-modal reasoning and alignment \citep{li_dual_2025, chen_unified_2026, rojas_diffuse_2025, hayakawa_mmdisco_2025}. Yet, for land-use scenario planning, the core unmet need is a multimodal generator that can jointly produce spatial layers with explicit cross-modal geometric consistency, rather than synthesizing modalities independently.

Second, current generative models lack sufficient controllability over planning objectives. In practice, urban planning is inherently goal-driven and therefore requires controlled generation: planners articulate explicit development objectives and constraints, and expect model outputs to respond predictably to these specified inputs. Diffusion models provide a strong foundation for controlled generation because they avoid common GAN limitations such as mode collapse and training instability \citep{karras_analyzing_2020, saxena_generative_2021}, and can integrate objectives as controls \citep{dhariwal_diffusion_2021, croitoru_diffusion_2023}. Technical advances in latent-space modeling \citep{rombach_high-resolution_2022}, optimized sampling strategies \citep{nichol_improved_2021}, and initialization refinements \citep{chung_come-closer-diffuse-faster_2022} have further improved their imagery generation efficiency. ControlNet marks an important milestone by injecting spatial conditions such as segmentation and depth maps into pretrained diffusion models without modifying their weights \citep{zhang_adding_2023}, while language--vision models such as CLIP and Imagen strengthen prompt grounding and prompt--output alignment \citep{radford_learning_2021, saharia_photorealistic_2022}. Multi-view and geometry-aware diffusion frameworks such as MVDream, 3D Adapter, and CMD further improve three-dimensional consistency and controllable editing across views \citep{shi_mvdream_2024, chen_3d-adapter_2025, li_cmd_2025}. Recent planning-oriented studies have also begun to operationalize diffusion-based generation with geospatial controls and iterative human guidance \citep{he_human-guided_2026, sun_envisioning_2026}. Despite this progress, the controllability in generative urban models remains limited: current control interfaces are predominantly generic rather than grounded in the context of land-use planning. 

Third, most of the generative outputs are evaluated by their general visual realism, but much less by domain-specific criteria such as land-use metrics. In existing generative studies, evaluation relies primarily on generic visual realism metrics such as Fréchet Inception Distance (FID), which assess perceptual fidelity but do not verify whether outputs satisfy domain-specific requirements. Visual realism alone cannot assess whether a generated land-use scenario is geographically valid, structurally consistent across modalities, or aligned with stated planning objectives. Although diffusion models have substantially improved output fidelity \citep{dhariwal_diffusion_2021, croitoru_diffusion_2023}, the evaluation of spatial generation remains challenging without explicit domain-specific benchmarks. This limitation makes verifiable evaluation a core requirement of our work: planning-oriented generative frameworks need explicit and interpretable verification, for cross-modal structural coherence and for objective compliance that go beyond generic image realism metrics.

\subsection{Scenario planning and evaluation}
Scenario planning is a common approach for exploring multiple plausible urban futures before a municipality commits to a specific development path. Rather than producing a single forecast, it constructs a set of scenarios that bound key uncertainties and make tradeoffs explicit for decision-making \citep{schoemaker_when_1991, van_notten_updated_2003}. In urban planning practice, scenario planning is particularly common in land-use, transportation and growth management contexts, where alternative development patterns are compared to support public communication and deliberation \citep{bartholomew_land_2007}. A typical workflow includes: (1) defining general principles; (2) constructing narratives and translating them into specific objectives such as zoning requirements, growth expectations, or land density targets; (3) generating spatial scenarios using sketch planning, or land-use simulation models; and (4) evaluating impacts through indicators or multi-criteria methods, often supported by participatory planning tools \citep{couclelis_where_2005, labiosa_integrated_2013}. In fact, such an iteration over principle definition, goal specification, scenario generation, and lastly evaluation is mimicked in our design of DualControlNet.

Predictive land-use models can support scenario planning by using land-use change dynamics to forecast future land-use transitions. However, the literature has identified several structural limitations of predictive modeling in exploratory and design-oriented planning contexts.
First, there is a high technical barrier and adoption friction in practice. Predictive models often require extensive data preparation, calibration, and parameter tuning, and the lack of institutional capacity and user familiarity can hinder their practical uptake \citep{vonk_bottlenecks_2005, vonk_improving_2008}. This technical complexity may limit their accessibility in iterative and participatory planning processes.
Second, many predictive models are explicitly calibrated to reproduce historical transition patterns. As noted in scenario planning literature, such calibration can bias model outputs toward path dependence and incremental change, making it difficult to represent disruptive futures or normative visions that intentionally depart from business-as-usual trajectories \citep{couclelis_where_2005, molinero-parejo_combining_2023}. Because the modeling logic is grounded in transition probabilities derived from past data, alternative spatial configurations that lack historical precedent may be underexplored.
Third, validation in predictive land-use modeling is commonly based on reproducing observed short-term spatial patterns. Reviews of land-use modeling highlight challenges related to non-stationarity, equifinality, and structural uncertainty when extending projections over longer horizons \citep{van_vliet_review_2016}. These characteristics complicate the interpretation of long-term scenario outputs, particularly when scenarios aim to explore transformative change.
Fourth, even within established modeling families such as urban cellular automata, considerable diversity exists in model structures and parameterizations. This heterogeneity complicates transparent model selection and explanation to stakeholders, potentially reducing interpretability in collaborative planning settings \citep{sante_cellular_2010}.
Finally, because no single predictive framework is universally applicable across contexts and policy questions, scenario planning often benefits from multi-model comparison approaches; however, such practices increase analytical workload and coordination complexity \citep{hewitt_theres_2022}. 
Together, these limitations motivate complementary approaches that enable faster iteration, clearer control of scenario assumptions, and more communicable outputs, while treating predictive land-use models as one component within a broader planning support workflow rather than the sole scenario generator \citep{couclelis_where_2005}.

\subsection{Summary of research landscape and gaps}
Traditional land-use models have provided valuable tools for quantifying and forecasting land development using interpretable land-use metrics. However, despite their methodological diversity, these approaches remain predictive: they rely on historical data to predict gradual transition in existing land-use categories, rather than directly creating new land-use maps. Generative AI, especially diffusion models, has made it feasible to synthesize high-fidelity urban representations, such as satellite imagery or building footprints \citep{ho2020denoising, rombach_high-resolution_2022, wu2022ganmapper, dedring2024synthesis, sun_envisioning_2026, he_human-guided_2026}. However, urban development practice still calls for a unified generative framework that supports multimodal joint generation, controllability based on targeted land-use metrics, and domain-specific evaluation beyond general visual realism \citep{lowry2014comparing, abou2022overview, castro2025comprehensive}.

\section{Data and Methods}
\label{sec:method}
\begin{figure}[H]
    \centering
    \includegraphics[width=\linewidth, keepaspectratio]{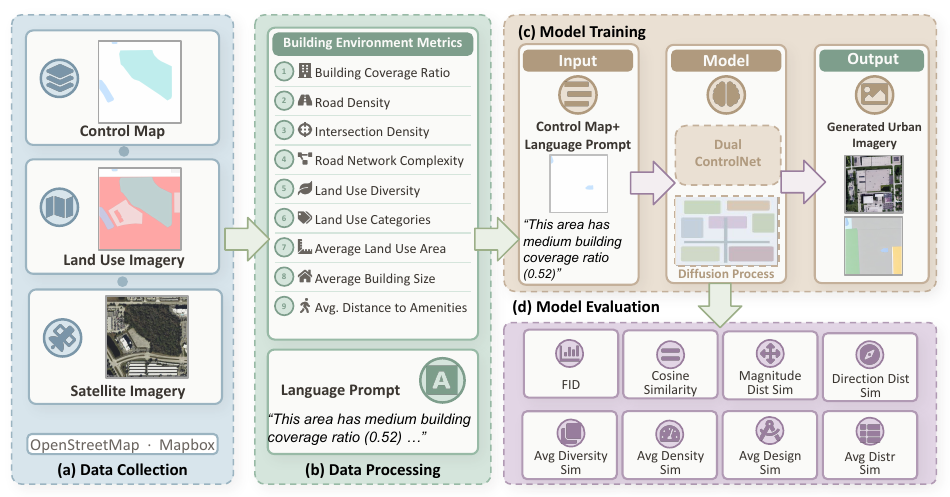}
    \caption{Workflow for joint land-use map and satellite imagery generation}
    \label{fig:multimodal}
\end{figure}

Figure~\ref{fig:multimodal} illustrates the overall workflow of the DualControlNet framework for land-use scenario generation and evaluation. The pipeline consists of four stages: (a) data collection, (b) data processing, (c) model training, and (d) model evaluation. In the data collection stage, we assemble multimodal urban data, including satellite imagery, land-use imagery, and control maps derived from OpenStreetMap and Mapbox. During data processing, we compute a set of land use metrics, such as building coverage ratio, road density, land-use diversity, and average land-use area, which are further translated into structured language prompts to guide generation. In the model training stage, DualControlNet takes both control maps and metric-informed language prompts as inputs to jointly generate land use and satellite imagery. Finally, DualControlNet is evaluated from three sets of metrics: domain-specific land-use metrics, spatial consistency between land-use maps and satellite imagery, and visual realism. 

\subsection{Data Collection and Processing}
\begin{figure}[H]
    \centering
    \includegraphics[width=\linewidth, keepaspectratio]{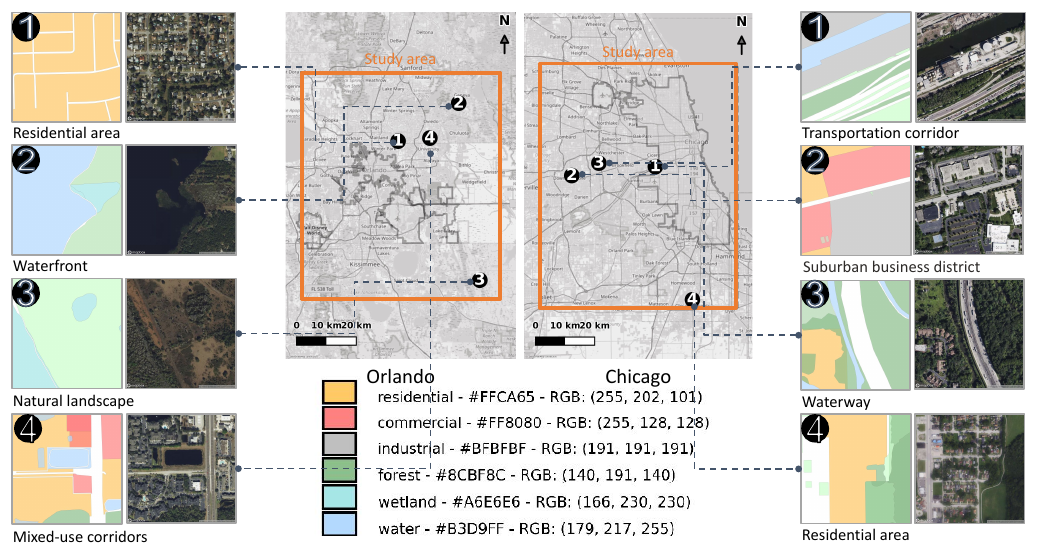}
    \caption{Study areas and examples of spatially aligned satellite and land use images}
    \label{fig:data}
\end{figure}

\begin{figure}[H]
    \centering
    \includegraphics[width=\linewidth, keepaspectratio]{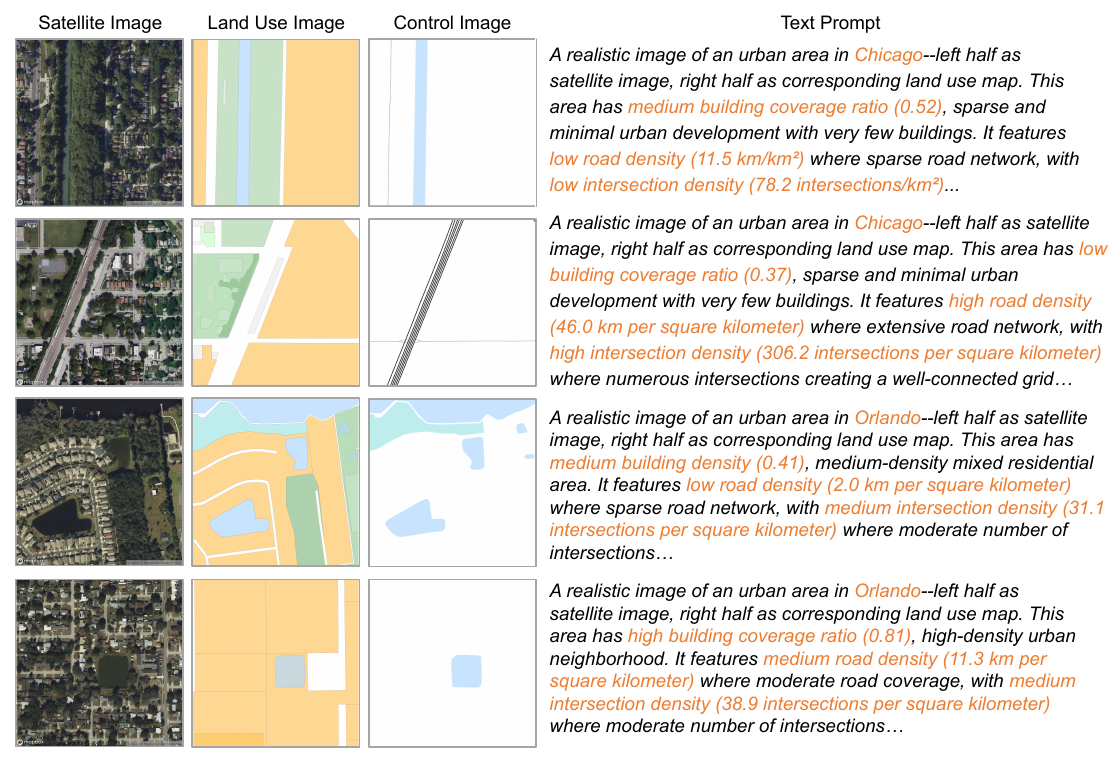}
    \caption{Example of multimodal data representations: text prompts, satellite images, land-use maps, and control images}
    \label{fig:datap}
\end{figure}

We first developed a multi-source data collection framework combining high-resolution satellite imagery and land use maps from OpenStreetMap. The dataset was collected using a custom-built GeoDataCollector that systematically samples urban areas through a grid-based approach. We collected satellite image data from two U.S. cities: Chicago and Orlando. The data was collected centered around each city's downtown area. For Chicago, the collection area was divided into a 180 × 120 grid, resulting in 21,600 image tiles. For Orlando, a 150 × 150 grid was used, yielding 22,500 image tiles. Each tile covers a fixed ground area of about 0.45 km × 0.45 km. The method calculates the exact latitude and longitude boundaries of each tile. Figure~\ref{fig:data} illustrates the study areas and provides examples of the collected satellite and land use images.
For satellite imagery, we used Mapbox Satellite v9, which provides high-resolution imagery primarily sourced from Maxar/DigitalGlobe between 2011-2023. Images were acquired through the Mapbox Static Images API at a resolution of 512x512 pixels, with consideration of geographical coordinates and zoom level 17 to maintain consistent spatial coverage across different latitude positions. Using OpenStreetMap, we gathered land use maps with diverse functions. Land use categories include residential, commercial, and industrial zones; urban service facilities such as educational institutions, healthcare centers, and recreational spaces; natural landscape elements such as forests, water bodies, and open green areas. These land-use categories together constitute the land use map. Lastly, we created a set of control maps to guide the generation of land use and satellite imagery. The control maps include water areas and transportation infrastructure. We extracted water areas from OpenStreetMap, including rivers, lakes, and other significant water bodies. As summarized in Figure~\ref{fig:datap}, we collected three types of spatially aligned images: (1) high-resolution satellite imagery showing the physical characteristics of the urban landscape, (2) detailed land use maps displaying the spatial organization of land-use elements, and (3) control maps featuring water areas and major road networks. 

Imagery alone cannot make generation controllable or verifiable, because a planner specifies a scenario in numbers rather than in pixels. We therefore compute nine quantitative metrics for every tile, grouped into two families that play distinct roles in our framework (Table~\ref{tab:3dmetric}). The first family, land-use metrics, is the target of both control and verification. Building coverage ratio measures how intensively land is developed; land-use diversity, computed as a Shannon entropy over land-use proportions, measures how functionally mixed it is; the number of land-use categories records how many distinct functions are present; and average land-use area measures the spatial granularity at which those functions are organized. These four quantities can be specified as a planning objective on the input side and recovered from a generated map on the output side, which allows a generated scenario to be verified.

The second family, general built environment metrics, describes the morphological and infrastructural context in which land use is embedded. Road density and intersection density together characterize the extent and granularity of the street network, road network complexity summarizes its structural intricacy as the ratio of segments to intersections, average building size records the typical footprint of individual structures, and average distance to amenities captures local accessibility to urban services. Unlike the land-use metrics, these quantities are not recoverable from a generated land-use map and hence do not serve as verification targets. Instead, they condition generation by situating each scenario within a plausible urban form, ensuring that a prompted level of development is realized with a street network and building fabric consistent with real cities. 

\begin{table}[H]
\centering
\renewcommand{\arraystretch}{1.15}
\caption{Summary of land-use and built environment metrics}
\label{tab:3dmetric}

\begin{tabularx}{\linewidth}{@{} p{4.4cm} X p{6.3cm} @{}}
\toprule
\textbf{Metric} & \textbf{Description} & \textbf{Formula} \\
\midrule

\multicolumn{3}{@{}l}{\textbf{Land Use Metrics}} \\
\midrule

Building Coverage Ratio &
Proportion of land area covered by buildings. &
$BCR = \frac{Area_{\text{buildings}}}{Area_{\text{total}}}$ \\

Land Use Diversity &
Shannon diversity index, where $p_i$ is the proportion of land area for land-use type $i$. &
$LUD = -\sum_{i=1}^{n} p_i \ln(p_i)$ \\

Land Use Categories &
Number of distinct land-use types in the area. &
$N_{\text{categories}}$ \\

Average Land Use Area &
Average area per land-use category, where $Area_i$ is the area of type $i$. &
$ALUA = \frac{\sum_{i=1}^{n} Area_i}{n}$ \\

\midrule
\multicolumn{3}{@{}l}{\textbf{Other Built Environment Metrics}} \\
\midrule

Road Density &
Total length of roads per square kilometer. &
$RD = \frac{Length_{\text{roads}}}{Area_{\text{km}^2}}$ \\

Intersection Density &
Number of road intersections per square kilometer. &
$ID = \frac{N_{\text{intersections}}}{Area_{\text{km}^2}}$ \\

Road Network Complexity &
Ratio of road segments (edges) to intersections (nodes). &
$RNC = \frac{N_{\text{edges}}}{N_{\text{nodes}}}$ \\

Average Building Size &
Average area per building, where $Area_j$ is the area of building $j$. &
$ABS = \frac{\sum_{j=1}^{m} Area_j}{m}$ \\

Average Distance to Amenities &
Mean distance from the center point to all amenities. &
$ADA = \frac{\sum_{i=1}^{n} d_i}{n}$ \\

\bottomrule
\end{tabularx}
\end{table}

The nine metrics in Table~\ref{tab:3dmetric} are numeric, whereas a diffusion model is conditioned through language. Our first processing step therefore translates each tile's numeric metrics into a structured text prompt, illustrated in the Text Prompt column of Figure~\ref{fig:datap}. The translation is deliberately hybrid: every metric enters the prompt twice, once as a qualitative descriptor and once as its exact numeric value in parentheses. A prompt reads, for example, ``\textit{A realistic image of an urban area in Chicago...This area has medium building coverage ratio (0.52), sparse and minimal urban development with very few buildings. It features low road density (11.5 km/km$^2$) where sparse road network, with low intersection density (78.2 intersections/km$^2$)...}'' Each prompt thus opens by naming the city, and then proceeds metric by metric through building coverage ratio, land-use diversity, road density, intersection density, and the remaining quantities, rendering each as a categorical level, followed by the measured value and a short verbal description of what that value means for urban form. The multimodal data representations in Figure~\ref{fig:datap} complete our data preparation for DualControlNet, as the control images and text prompts serving as inputs and the paired satellite imagery and land-use maps as outputs. 

\subsection{Diffusion Models and ControlNet Baselines}
We adopt Stable Diffusion as the baseline model. Diffusion models generate samples by learning a reverse denoising process that inverts a forward process, in which Gaussian noise is gradually added to a clean latent sample $z_0$ to obtain a noisy latent $z_t$ at timestep $t$. Given a conditioning input $x$, which is the text prompt in our setting, a neural network $\epsilon_{\theta}$ is trained to predict the injected noise $\epsilon$ at each timestep. The training objective is:
\begin{equation}
L_{(\theta)} = \mathbb{E}_{z_0, t, \epsilon \sim \mathcal{N}(0,1)}\left[ \| \epsilon - \epsilon_{\theta}(z_t, t, \tau_{\theta}(x)) \|^2 \right].    
\label{eq:loss_training}
\end{equation}

Although text prompts enable diverse image synthesis, they provide limited spatial controllability for land-use map generation. We therefore build on ControlNet \citep{zhang_adding_2023}, which augments a pretrained diffusion U-Net with additional conditioning while leaving the original pretrained weights intact. ControlNet duplicates the base U-Net into a frozen copy and a trainable copy, and introduces two zero-initialized $1 \times 1$ convolutions, denoted $\mathcal{Z}(\cdot;\cdot)$, to inject the control signal and merge the conditional residual. Let $\mathcal{F}(\cdot; \Theta)$ denote the frozen base model and $\mathcal{F}(\cdot; \Theta_c)$ the trainable copy, initialized with $\Theta_c=\Theta$. Given a control condition $c$, the conditioned output is:
\begin{equation}
y_c = \mathcal{F}(x; \Theta) + \mathcal{Z}\!\left(\mathcal{F}\!\left(x + \mathcal{Z}(c; \Theta_{z1}); \Theta_c\right); \Theta_{z2}\right),   
\end{equation}
and the denoising objective for training the ControlNet branch is:
\begin{equation}
 L_{(\Theta_c, \Theta_{z1}, \Theta_{z2})} = \mathbb{E}_{z_0, t, c, \epsilon \sim \mathcal{N}(0,1)} \left[ \| \epsilon - \epsilon_{\Theta_c, \Theta_{z1}, \Theta_{z2}}(z_t, t, c) \|^2 \right].   
\end{equation}
\noindent This objective is consistent with standard diffusion training practice \citep{dhariwal_diffusion_2021, croitoru_diffusion_2023}. Because the control signal enters through the trainable copy alone, the formulation lets generation follow the explicit spatial constraints carried by the control images, which is essential for our tasks that demand spatial structures.

\subsection{Designing DualControlNet}
\begin{figure}[ht!]
    \centering
    \includegraphics[width=\linewidth,keepaspectratio]{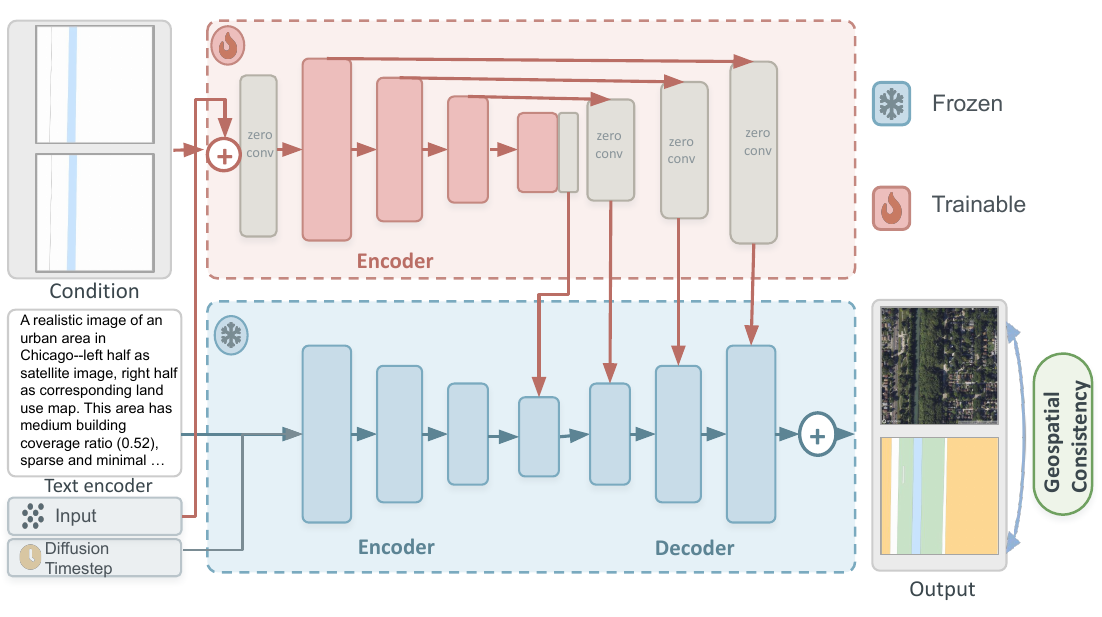}
    \caption{Demonstration of DualControlNet}
    \label{fig:archi}
\end{figure}
Multimodal land-use scenario planning needs not only realistic imagery, but also cross-modal consistency: roads, blocks, and parcel boundaries should align between the satellite imagery and the land-use maps. In practice, training two separate ControlNets for satellite imagery and land-use maps can produce individually plausible outputs, yet their geometric structures may drift because the two outputs are optimized independently and do not share an explicit mechanism to enforce spatial alignment. 

Therefore, we propose DualControlNet to support multimodal generation with cross-modal spatial consistency. The core idea is to extend the ControlNet formulation to a dual-modality setting and introduce a geospatial consistency loss that regularizes the geometric alignment between the generated land-use map and satellite imagery, as illustrated in Figure~\ref{fig:archi}. DualControlNet is not only a higher-fidelity generator, but also a multimodal generator whose outputs are spatially aligned and more amenable to evaluation against land-use metrics. This modification is implemented within the training objective computed at each diffusion timestep $t$. 

Given a clean latent $\mathbf{x}_0 \in \mathbb{R}^{B \times C \times H \times W}$, the forward process samples noise $\boldsymbol{\epsilon}\sim\mathcal{N}(0,\mathbf{I})$ and produces $\mathbf{x}_t \sim q(\mathbf{x}_t\mid \mathbf{x}_0)$. The network predicts a target $\mathbf{y}_{\text{target}}$ under a chosen parameterization (noise, $\mathbf{x}_0$, or velocity). Under velocity prediction,
\begin{equation}
 \mathbf{v} = \alpha_t \boldsymbol{\epsilon} + \beta_t \mathbf{x}_0,
\end{equation}
where $\alpha_t$ and $\beta_t$ are determined by the noise schedule.
We optimize the denoising objective
\begin{equation}
\mathcal{L}_{\text{simple}} = \mathbb{E}_{t, \mathbf{x}_0, \boldsymbol{\epsilon}} \left[ \left\| \hat{\mathbf{y}} - \mathbf{y}_{\text{target}} \right\|_2^2 \right],
\end{equation}
Following standard practice, we use log-variance reweighting and an additional VLB term, which are combined into the overall training objective $\mathcal{L}_{\text{total}}$ \citep{dhariwal_diffusion_2021, croitoru_diffusion_2023}.
\begin{equation}
\mathcal{L}_{\text{total}} = \lambda_{\text{simple}} \cdot \mathcal{L}_{\text{reweighted}} + \lambda_{\text{vlb}} \cdot \mathcal{L}_{\text{vlb}}.
\end{equation}
\noindent These formulations follow standard diffusion training practice \citep{dhariwal_diffusion_2021, croitoru_diffusion_2023}, and the conditioning mechanism is adapted from ControlNet \citep{zhang_adding_2023}.

Our key modification is to add a geospatial consistency loss computed on the reverse-predicted clean latent $\mathbf{x}_0^{\text{pred}}$, reconstructed from $\mathbf{x}_t$ and the model output $\hat{\mathbf{y}}$:
\begin{equation}
\mathbf{x}_0^{\text{pred}} = \frac{\mathbf{x}_t - \sqrt{1 - \bar{\alpha}_t} \cdot \hat{\mathbf{y}}}{\sqrt{\bar{\alpha}_t}},   
\end{equation}
where $\bar{\alpha}_t$ is the cumulative product of the noise schedule up to timestep $t$.
We split $\mathbf{x}_0^{\text{pred}}$ along the width dimension at the midpoint $W_{\text{mid}} = \lfloor W / 2 \rfloor$:
\begin{equation}
\mathbf{x}_{0,\text{left}}^{\text{pred}} = \mathbf{x}_0^{\text{pred}}[:, :, :, :W_{\text{mid}}], \quad
\mathbf{x}_{0,\text{right}}^{\text{pred}} = \mathbf{x}_0^{\text{pred}}[:, :, :, W_{\text{mid}}:].
\end{equation}
To capture geometric boundaries, we compute finite-difference gradients along height and width:
\begin{equation}
\nabla_h^{\text{left}} = \mathbf{x}_{0,\text{left}}^{\text{pred}}[:, :, 1:, :] - \mathbf{x}_{0,\text{left}}^{\text{pred}}[:, :, :-1, :], \quad
\nabla_h^{\text{right}} = \mathbf{x}_{0,\text{right}}^{\text{pred}}[:, :, 1:, :] - \mathbf{x}_{0,\text{right}}^{\text{pred}}[:, :, :-1, :].    
\end{equation}
\begin{equation}
\nabla_w^{\text{left}} = \mathbf{x}_{0,\text{left}}^{\text{pred}}[:, :, :, 1:] - \mathbf{x}_{0,\text{left}}^{\text{pred}}[:, :, :, :-1], \quad
\nabla_w^{\text{right}} = \mathbf{x}_{0,\text{right}}^{\text{pred}}[:, :, :, 1:] - \mathbf{x}_{0,\text{right}}^{\text{pred}}[:, :, :, :-1].
\end{equation}
To address possible size mismatch, we crop both sides to the minimum overlapping size:
\begin{equation}
H_{\min} = \min(\text{height}(\nabla_h^{\text{left}}), \text{height}(\nabla_h^{\text{right}})), \quad
W_{\min}^h = \min(\text{width}(\nabla_h^{\text{left}}), \text{width}(\nabla_h^{\text{right}})),
\end{equation}
\begin{equation}
H_{\min}^w = \min(\text{height}(\nabla_w^{\text{left}}), \text{height}(\nabla_w^{\text{right}})), \quad
W_{\min} = \min(\text{width}(\nabla_w^{\text{left}}), \text{width}(\nabla_w^{\text{right}})).
\end{equation}
We define the contour consistency loss as the sum of L1 distances between the cropped gradient maps:
\begin{equation}
\mathcal{L}_{\text{geo}}^{(h)} =
\text{l1}(\nabla_h^{\text{left}}[:, :, :H_{\min}, :W_{\min}^h], \nabla_h^{\text{right}}[:, :, :H_{\min}, :W_{\min}^h]),
\end{equation}
\begin{equation}
\mathcal{L}_{\text{geo}}^{(w)} =
\text{l1}(\nabla_w^{\text{left}}[:, :, :H_{\min}^w, :W_{\min}], \nabla_w^{\text{right}}[:, :, :H_{\min}^w, :W_{\min}]),
\end{equation}
\begin{equation}
\mathcal{L}_{\text{geo}} = \mathcal{L}_{\text{geo}}^{(h)} + \mathcal{L}_{\text{geo}}^{(w)}.
\end{equation}
This auxiliary term encourages the two modalities to share consistent boundary geometry, improving cross-modal structural alignment. We apply a small weight $\lambda_{\text{geo}}$ to avoid overpowering the denoising objective. The final objective is:
\begin{equation}
\mathcal{L}_{\text{final}} = \mathcal{L}_{\text{simple}} + \lambda_{\text{vlb}} \cdot \mathcal{L}_{\text{vlb}} + \lambda_{\text{geo}} \cdot \mathcal{L}_{\text{geo}}.
\end{equation}

In our experiments, we set $\lambda_{\text{geo}} = 0.001$ to balance geometric regularization and generative performance. By integrating this geospatial prior into the latent diffusion framework, DualControlNet improves cross-modal structural coherence while retaining the high-fidelity synthesis of diffusion models.

\section{Experiment Design}
\label{sec:experiment}

\subsection{Model evaluation}
A generated land-use map is useful to planning only if it is verifiable, and verification demands more than visual plausibility. We therefore evaluate the models along three groups of metrics, summarized in Table~\ref{tab:metric_explanations1}. The first group, land-use metrics, asks whether a generated map satisfies the planning objectives that the prompt specified, comparing the building coverage ratio, land-use diversity, average land-use area, and land-use composition of the generated map against the real maps. The second group, spatial consistency metrics, asks whether the geometric structures in the satellite image and the land-use map align. The third group, visual realism, asks whether the generated satellite imagery is perceptually credible. The three groups are complementary: a model can produce a realistic satellite image that contradicts its own land-use map, or a pair of well-aligned images that ignores the requested density. Because DualControlNet generates the paired modalities jointly, we can quantify all three groups without relying on subjective inspection.

\begin{table}[ht!]
\centering
\renewcommand{\arraystretch}{1.15}
\caption{Eight evaluation metrics used in model assessment.}
\label{tab:metric_explanations1}

\begin{tabularx}{\linewidth}{@{} p{3.4cm} X p{6.5cm} @{}}
\toprule
\textbf{Metric} & \textbf{Description} & \textbf{Formula} \\
\midrule

\addlinespace[2pt]
\multicolumn{3}{@{}l@{}}{\textbf{Land use metrics}} \\
\addlinespace[2pt]

Avg. Land Use Diversity Consistency &
Measures similarity in land-use diversity between generated and reference maps. &
$S_{\mathrm{LUD}} 
= 1 - 
\frac{\left| LUD_g - LUD_r \right|}
{\max(LUD_g, LUD_r)}$ \\

Avg. Building Coverage Ratio Consistency &
Measures similarity in building coverage ratio between generated and reference land-use maps. &
$S_{\mathrm{BCR}} 
= 1 - 
\left| BCR_g - BCR_r \right|$ \\

Avg. Average Land Use Area Consistency &
Measures similarity in average land-use area between generated and reference land-use maps. &
$S_{\mathrm{ALUA}} 
= 1 - 
\frac{\left| ALUA_g - ALUA_r \right|}
{\max(ALUA_g, ALUA_r)}$ \\

Avg. Distribution Similarity &
Evaluates statistical similarity of land-use proportion distributions between generated and reference maps. &
$S_{\text{dist}}
= 1 - 
\frac{1}{n}
\sum_{i=1}^{n}
\left| p_i - q_i \right|$ \\
\midrule

\addlinespace[2pt]
\multicolumn{3}{@{}l@{}}{\textbf{Spatial consistency}} \\
\addlinespace[2pt]
Cosine Similarity &
Measures directional alignment between generated remote sensing imagery and the corresponding land-use map, reflecting global orientation consistency. &
$\text{Cosine Similarity}=\frac{\sum_{i=1}^{N} G_1(i)\,G_2(i)}{\sqrt{\sum_{i=1}^{N} G_1(i)^2}\,\sqrt{\sum_{i=1}^{N} G_2(i)^2}}$ \\

Magnitude Distribution Similarity &
Compares the distribution of vector magnitudes between the generated satellite image and land-use image. &
$S_{\text{mag}}=\frac{1}{1+W(P_1^m,P_2^m)}$ \\

Direction Distribution Similarity &
Evaluates whether directional features are preserved between the generated satellite image and land-use image. &
$S_{\text{dir}}=1-d_{\text{cos}}(P_1^d,P_2^d)$ \\
\midrule

\addlinespace[2pt]
\multicolumn{3}{@{}l@{}}{\textbf{Visual realism}} \\
\addlinespace[2pt]
Fréchet Inception Distance (FID) &
Compares generated remote sensing imagery with real images to quantify overall realism and generation quality. &
$\mathrm{FID} = \|\mu_r - \mu_g\|^2 + \mathrm{Tr}\!\left(\Sigma_r + \Sigma_g - 2(\Sigma_r \Sigma_g)^{1/2}\right)$ \\

\bottomrule
\end{tabularx}
\end{table}

The land-use metrics are the core of our verification, and computing them requires the post-processing of generated land-use maps. The model returns a land-use map as an image rather than as a vector layer, so the same metrics we defined in Table~\ref{tab:3dmetric} must be recovered from the pixels. We recover them as follows: for land-use diversity and the number of land-use categories, the generated map is downsampled and converted into an RGB pixel array, and color quantization groups similar colors into clusters; we retain only clusters occupying more than 1\% of the image and discard very bright colors, which correspond to unclassified areas. The retained clusters provide the set of land-use categories, and their area shares provide the land-use composition. For building coverage ratio, we identify building pixels through predefined RGB ranges, construct a binary mask, and take the ratio of building pixels to total pixels. For average land-use area, we build a binary mask per retained category, apply connected-component labeling to delineate contiguous parcels, discard components below a minimum-size threshold, and average the areas of the remainder. Each of the four land-use metrics is then converted into a consistency score, defined in Table~\ref{tab:metric_explanations1}, such that a higher value indicates that the generated map reproduces the reference map and a lower value indicates larger differences. 

The spatial consistency metrics evaluate whether the satellite image and the land-use map depict the same urban form. We assess it through image gradients, since parcel boundaries, road edges, and block faces can be approximated by gradients in both land-use and satellite images. Cosine similarity compares the direction of the two gradient fields, magnitude distribution similarity compares their distributions of gradient strength, and direction distribution similarity compares their distributions of gradient orientation. Visual realism is assessed with the Fréchet Inception Distance (FID), which compares latent distributions of generated and real satellite imagery, with lower values indicating closer agreement. FID measures only perceptual credibility and not geographic validity.

\subsection{Experiment process}
All models were trained and evaluated on high-performance computing platform, using both NVIDIA A100 and L4 GPUs. The dataset comprises over 40,000 paired satellite and land use images from Chicago and Orlando for training, with an additional 4,000 pairs reserved for testing. Four different models were developed and compared, each trained for 5 epochs, with training for each model taking approximately 20 GPU hours. Four distinct model configurations are compared. The first two models are built upon our DualControlNet architecture. The first model, DualControlNet without (w/o) Geo Loss, serves as a baseline. While capable of dual-task generation, it deliberately omits our proposed geospatial loss function. In contrast, the second model, DualControlNet shares the same architecture but crucially integrates the geospatial loss to enhance the spatial consistency between the generated satellite and land use outputs. The third model, ControlNet (Land use), is specialized to generate only the land use map from a single input control map. The fourth model, ControlNet (Satellite), is dedicated solely to generate the satellite image from its respective control map.

\section{Results}
\label{sec:results}
\subsection{Evaluating model performance}
Table~\ref{tab:model_performance_2} and Table~\ref{tab:model_performance_1} compare four models across land-use, spatial consistency, and visual realism metrics in Chicago and Orlando. Two of these models are unimodal ControlNet baselines, conditioned on land use or satellite imagery alone, and two are DualControlNet variants that jointly generate both modalities, with and without the geospatial loss. Our comparison therefore proceeds in two steps: we first contrast multimodal DualControlNet against the unimodal baselines to isolate the value of controllable and verifiable generation, and then contrast DualControlNet with and without the geospatial loss to isolate the loss's contribution to joint generation. We begin with land-use metrics, as these indicators directly reflect whether the generated outputs align with the text prompts and can be verified against domain-specific measures. We then turn to spatial consistency and visual realism metrics, with the former tackling multimodal coherence and the latter serving as a general evaluation of imagery quality.

Both DualControlNet variants substantially outperform the unimodal ControlNet in terms of aligning land-use metrics. Relative to the unimodal baselines, the best DualControlNet improves Land Use Diversity Consistency from 0.3428 to 0.5580, Average Land Use Area Consistency from 0.1755 to 0.2624, Building Coverage Ratio Consistency from 0.6362 to 0.7400, and Distribution Similarity from 0.8249 to 0.8734. Such increases correspond to percentage gains of 62.8\%, 49.5\%, 16.3\%, and 5.9\%. The improvement shows that jointly generating land use together with satellite imagery, rather than generating land use or satellite imagery in isolation, yields outputs that are both more controllable and more faithful to the target land-use metrics. DualControlNet with geospatial loss attains the highest Average Building Coverage Ratio Consistency of 0.7400 and the highest Average Land Use Area Consistency of 0.2624, improving on the version without it by 5.8\% and 8.4\%, respectively, particularly in reconstructing density and building blocks. The version without geospatial loss retains a modest edge in Land Use Diversity Consistency (0.5580 vs. 0.4992, 11.8\%) and Distribution Similarity (0.8734 vs. 0.8570, 1.9\%), suggesting that relaxing the geospatial constraint allows slightly greater flexibility in learning diverse land-use compositions. Importantly, the satellite-only ControlNet produces satellite imagery but no land-use map, so it cannot be verified on any land-use metric and is reported as N/A throughout Table~\ref{tab:model_performance_2}. This result again highlights verifiable generation as an important contribution of DualControlNet. 

\begin{table}[H]
\centering
\caption{Models evaluated by land-use metrics (I)}
\label{tab:model_performance_2}
\renewcommand{\arraystretch}{1.2}
\resizebox{.9\linewidth}{!}{%
\begin{tabularx}{\linewidth}{l*{4}{>{\centering\arraybackslash}X}}
\toprule
\textbf{Metric (City)} & \textbf{Dual\allowbreak Control\allowbreak Net w/o Geo Loss} & \textbf{Dual\allowbreak Control\allowbreak Net w/ Geo Loss} & \textbf{Control\allowbreak Net LandUse} & \textbf{Control\allowbreak Net Satellite} \\
\midrule
Avg. Land Use Diversity Con. (All) & \textbf{0.5580} & 0.4992 & 0.3428 & N/A \\
Avg. Land Use Diversity Con. (Orlando) & \textbf{0.5075} & 0.4391 & 0.3296 & N/A \\
Avg. Land Use Diversity Con. (Chicago) & \textbf{0.6084} & 0.5592 & 0.3560 & N/A \\
\addlinespace
Avg. Building Coverage Ratio Con. (All) & 0.6993 & \textbf{0.7400} & 0.6362 & N/A \\
Avg. Building Coverage Ratio Con. (Orlando) & 0.7805 & \textbf{0.7971} & 0.7323 & N/A \\
Avg. Building Coverage Ratio Con. (Chicago) & 0.6180 & \textbf{0.6829} & 0.5400 & N/A \\
\addlinespace
Avg. Average Land Use Area Con. (All) & 0.2421 & \textbf{0.2624} & 0.1755 & N/A \\
Avg. Average Land Use Area Con. (Orlando) & 0.2201 & \textbf{0.2337} & 0.1734 & N/A \\
Avg. Average Land Use Area Con. (Chicago) & 0.2640 & \textbf{0.2911} & 0.1775 & N/A \\
\addlinespace
Avg. Distribution Sim. (All) & \textbf{0.8734} & 0.8570 & 0.8249 & N/A \\
Avg. Distribution Sim. (Orlando) & \textbf{0.8711} & 0.8474 & 0.8208 & N/A \\
Avg. Distribution Sim. (Chicago) & \textbf{0.8756} & 0.8665 & 0.8290 & N/A \\
\bottomrule
\end{tabularx}
}
\end{table}

\begin{table}[H]
\centering
\caption{Models evaluated by spatial consistency and visual realism (II)}
\label{tab:model_performance_1}
\renewcommand{\arraystretch}{1.2}
\resizebox{.9\linewidth}{!}{
\begin{tabularx}{\linewidth}{l*{4}{>{\centering\arraybackslash}X}}
\toprule
\textbf{Metric (City)} & \textbf{DualControl\allowbreak Net w/o Geo Loss} & \textbf{DualControl\allowbreak Net w/ Geo Loss} & \textbf{ControlNet LandUse} & \textbf{ControlNet Satellite} \\
\midrule
FID (All)        & 61   & \textbf{58} & 213   & 61  \\
FID (Orlando)    & \textbf{76}   & 78     & 215 & 81  \\
FID (Chicago)    & 93   & \textbf{82}     & 248    & 89     \\

\addlinespace
Cosine Sim. (All) & 0.2180 & \textbf{0.2364} & 0.1118 &0.1083  \\
Cosine Sim. (Orlando) & 0.2473 & \textbf{0.2754} & 0.1122 & 0.1125  \\
Cosine Sim. (Chicago) & 0.1887 & \textbf{0.1973} & 0.1113 & 0.1041  \\
\addlinespace
Magnitude Dist. Sim. (All) & 0.9782 & \textbf{0.9804} & 0.9713 & 0.9733  \\
Magnitude Dist. Sim. (Orlando) & 0.9801 & \textbf{0.9817} & 0.9716 &0.9738  \\
Magnitude Dist. Sim. (Chicago) & 0.9763& \textbf{0.9791} & 0.9710 &0.9728  \\
\addlinespace
Direction Dist. Sim. (All) & 0.5415 & \textbf{0.5911} & 0.2894 & 0.3181  \\
Direction Dist. Sim. (Orlando) & 0.4757 &\textbf{0.5177} &0.2989 &0.3281 \\
Direction Dist. Sim. (Chicago) & 0.6072 & \textbf{0.6645}& 0.2799 &0.3081 \\
\bottomrule
\end{tabularx}
}
\end{table}

For spatial consistency metrics, both DualControlNet variants again dominate the unimodal baselines. The best DualControlNet more than doubles Cosine Similarity, reaching 0.2364 versus 0.1118 for ControlNet LandUse and 0.1083 for ControlNet Satellite, and roughly doubles Direction Distribution Similarity, reaching 0.5911 versus 0.2894 and 0.3181 respectively. Within the DualControlNet family, the geospatial loss then provides a consistent additional gain. Cosine Similarity increases from 0.2180 to 0.2364, an 8.4\% improvement; Magnitude Distribution Similarity rises from 0.9782 to 0.9804; and Direction Distribution Similarity increases substantially with a 9.2\% improvement. These gains confirm that the geospatial consistency loss enhances global and directional structural coherence across modalities. Lastly, we assess visual realism using FID scores. DualControlNet is far sharper than the land-use-only baseline, lowering FID from 213 to 58, and it performs on par with or better than the satellite-only baseline, which reaches 61. Adding the geospatial loss lowers the overall FID from 61 to 58, and in Chicago from 93 to 82, while slightly worsening FID in Orlando. These results indicate that multimodal joint generation, reinforced by geospatial constraints, not only improves cross-modal consistency but also enhances perceptual image quality.

Across Orlando and Chicago, performance is generally consistent for land-use metrics and spatial consistency. DualControlNet with geospatial loss achieves higher Building Coverage Ratio Consistency in Orlando (0.7971) than in Chicago (0.6829), while spatial similarity metrics are also slightly stronger in Orlando. However, Direction Distribution Similarity is higher in Chicago (0.6645), likely due to its denser and more regular street network, which strengthens directional alignment under control-map conditioning.

Overall, the model evaluation highlights that land-use metrics successfully provide domain-specific verification of generated land-use maps, while spatial consistency metrics indicate the importance of multimodal generation across urban layers. DualControlNet with geospatial loss achieves the best balance across these three dimensions, supporting multimodal generation, metric-based control, and domain-specific verification.

\begin{figure}[ht!]
    \centering
    \includegraphics[width=\linewidth, keepaspectratio]{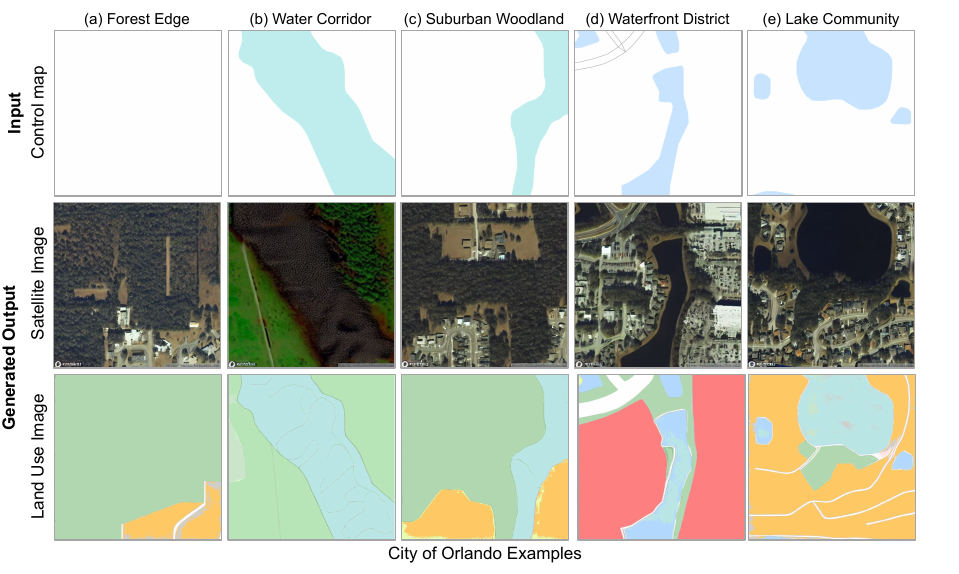}
    \caption{Examples of synthetic satellite images and land use images in Orlando}
    \label{fig:orex}
\end{figure}

\subsection{Generative baselines: reconstructing land-use patterns with original prompts}
Given a control map encoding only major roads and water bodies, DualControlNet can reconstruct a high-resolution satellite image and a land-use map corresponding to the original land-use prompts. Figure~\ref{fig:orex} illustrates synthetic land-use maps in Orlando across five diverse urban environments, including Forest Edge, Water Corridor, Suburban Woodland, Waterfront District, and Lake Community. 

Across these urban environment types, the generated land-use maps and satellite imagery preserve key structural elements from the control input. For example, in the Water Corridor and Lake Community cases, the model maintains the geometry of water bodies while generating realistic surrounding residential and green land uses. In the Forest Edge and Suburban Woodland examples, natural landscape patterns are reconstructed with gradual transitions between built and vegetated areas. In the Waterfront District scenario, the model captures dense development along the shoreline, aligning building footprints in the satellite image with corresponding urban land-use categories. The model also learns semantic correspondences between roof morphology in satellite imagery and land-use categories. For instance, single-family houses in Suburban Woodland and Lake Community scenarios exhibit smaller, detached roof structures associated with residential land use, whereas larger and more regular building footprints in Waterfront District areas correspond to commercial or mixed-use classifications.

\begin{figure}[H]
    \centering
    \includegraphics[width=\linewidth, keepaspectratio]{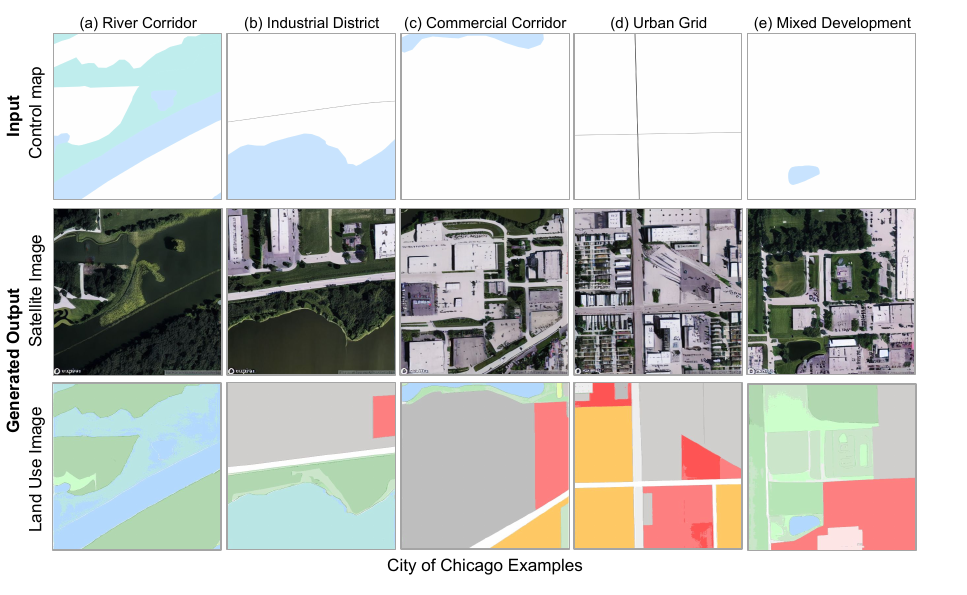}
    \caption{Examples of synthetic satellite images and land use images in Chicago}
    \label{fig:chex}
\end{figure}

Figure~\ref{fig:chex} demonstrates that DualControlNet can generate land-use maps for diverse urban environments similarly in Chicago, including River Corridor, Industrial District, Commercial Corridor, Urban Grid, and Mixed Development areas. Similar to Orlando, the model preserves the primary water and road structures specified in the control maps, demonstrating consistent generation aligning with imagery inputs. In the River Corridor scenario, DualControlNet clearly maintains the alignment between river geometry and adjacent green or industrial land use. In Industrial District and Commercial Corridor areas, the model generates larger building blocks and structured road networks consistent with high-intensity development. The Urban Grid example highlights the model’s ability to reconstruct regular block patterns and orthogonal street layouts characteristic of Chicago’s dense street network. In the Mixed Development case, diverse land-use categories are arranged at the block level, reflecting fine-grained spatial organization.

Compared to Orlando, Chicago examples exhibit higher building density and more regular grid structures, which are well captured in both satellite imagery and land-use maps. However, the visual realism of satellite imagery in Chicago is slightly lower (FID = 82) than in Orlando (FID = 78), with some brightness shifts and minor distortions in areas containing tall buildings and cast shadows. This difference may be attributed to the higher structural complexity and denser urban fabric in Chicago. 

\subsection{Generative scenario planning with single land-use dimensions}
\begin{figure}[ht!]
    \centering
    \includegraphics[width=0.8\linewidth, keepaspectratio]{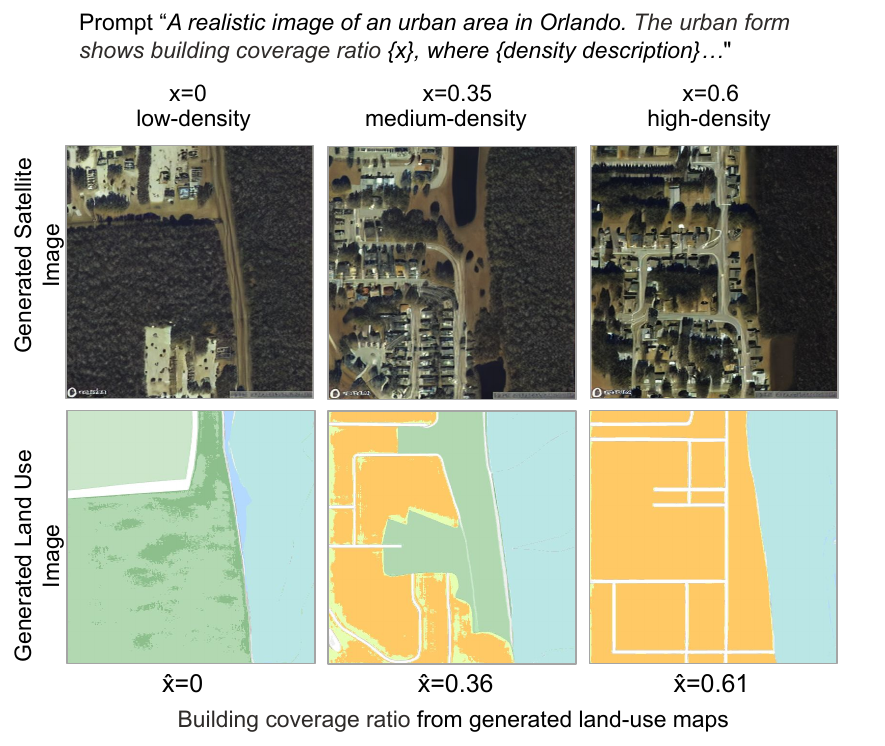}
    \caption{Generating land use and satellite imagery for various building coverage ratios}
    \label{fig:sen}
\end{figure}

To demonstrate the potential of generative scenario planning, we conduct a controlled experiment using a single land-use dimension, namely building coverage ratio. This dimension is chosen because building coverage ratio indicates the intensity of land development, which approximates sustainability, compactness, and even urban innovations. Here we use both language prompts that explicitly specify the target density level as ``low-density'', ``medium-density'', or ``high-density development'', together with numeric targets across three scenarios in which the building coverage ratio is set to be 0, 0.35, and 0.6. As shown in Figure~\ref{fig:sen}, the top row shows the generated satellite imagery, and the bottom row presents the corresponding land-use maps together with the building coverage ratio computed from the generated outputs.

The results illustrate the success of such uni-dimensional generative scenario planning since the land-use maps do respond to various density specifications. In the low-density scenario, where the building coverage ratio is set to 0, no residential buildings are generated. The land-use map is dominated by vegetated and open areas, and the satellite imagery reflects continuous green surfaces with minimal built structures. Roads remain sparse and primarily define the outer boundary of the site rather than organizing internal blocks. In the medium-density scenario, where the building coverage ratio is set to 0.35, residential buildings begin to cluster along internal road segments. Green areas are reduced and redistributed between building parcels, forming a balanced configuration between developed and open spaces. The road network becomes more structured and subdivides the site into smaller plots, supporting a recognizable neighborhood layout. The measured building coverage ratio from the generated land-use map is 0.36, closely matching the target value specified in the prompt. In the high-density scenario, where the building coverage ratio is set to 0.6, most land is occupied by residential buildings. Green areas are compressed into narrow buffers and residual patches. Roads assume a stronger organizing function, dividing the site into compact blocks that accommodate dense building footprints. The evaluated building coverage ratio reaches 0.61, again closely aligned with the target density. Across all three density levels, the difference between the specified and measured building coverage ratio remains minimal.

\begin{figure}[ht!]
    \centering
    \includegraphics[width=0.9\linewidth, keepaspectratio]{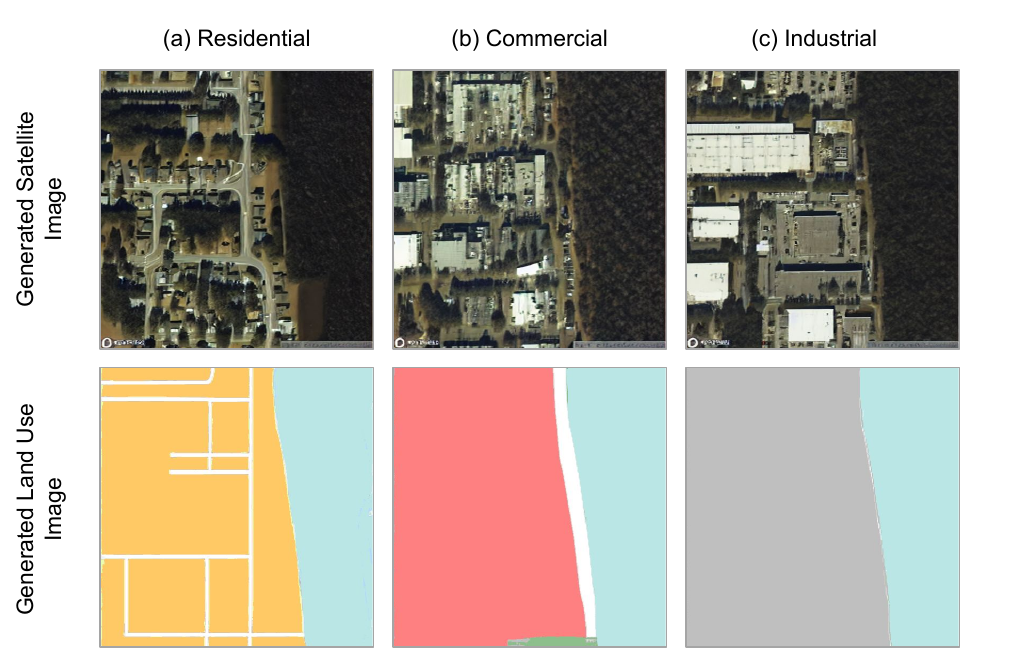}
    \caption{Diverse generated land-use maps and satellite images under the same prompts.}
    \label{fig:alt}
\end{figure}

With such uni-dimensional control, we also observe high randomness and diversity in generated outcomes. Using the same prompt, DualControlNet can produce multiple land-use maps and satellite images that satisfy the same density condition, and yet exhibit substantially different spatial and functional configurations. As shown in Figure~\ref{fig:alt}, DualControlNet produces three visually and functionally differentiated scenarios while maintaining the same building coverage ratio. In Figure~\ref{fig:alt}a, the generated satellite image exhibits smaller buildings with detached layouts and roof colors typical of single-family housing. The corresponding land-use map reflects this morphology through spatially distributed residential parcels. In Figure~\ref{fig:alt}b, the built structures become moderately sized with more varied shapes and roof tones, forming a denser and more heterogeneous configuration consistent with commercial development. In Figure~\ref{fig:alt}c, the built environment is dominated by large and simple building footprints with uniform roofs, visually resembling industrial facilities. In each case, the satellite imagery and land use map maintain cross-modal coherence, while the generated outputs reflect distinct functional identities. 

This capability contrasts with traditional predictive land use models, which operate primarily on numeric transitions of land use categories and cannot generate highly diverse visual representations. Predictive models can estimate how much land may convert from one category to another, but they do not synthesize the corresponding building forms, roof structures, or spatial textures visible in remote sensing imagery, let alone the highly diverse visual typologies that may emerge from simple textual prompts. In contrast, our generative land-use modeling framework simultaneously produces satellite imagery and land use maps, conditioning on both quantitative density specifications and qualitative descriptions.

\subsection{Generative scenario planning with multiple land-use dimensions}
Planning practice rarely pursues a single land-use objective in isolation. For example, a density target is typically pursued together with a diversity target, since compact development without a mix of functions produces neither walkable neighborhoods nor local access to jobs and services. Generative scenario planning is therefore useful to planners only if it can respond to several land-use dimensions at once. To test this capability, we jointly control building coverage ratio and the number of land use categories to simulate a sustainable development scenario. We deliberately select relatively underdeveloped sites in both Chicago and Orlando: sites with near-zero built coverage and only two or three land use categories in the ground truth. These sites provide the largest headroom for testing whether the model can generate a sustainable development scenario under multidimensional land-use controls, namely one with higher land-use density and higher land-use diversity than the current condition.

\begin{figure}[ht!]
    \centering
    \includegraphics[width=0.9\linewidth, keepaspectratio]{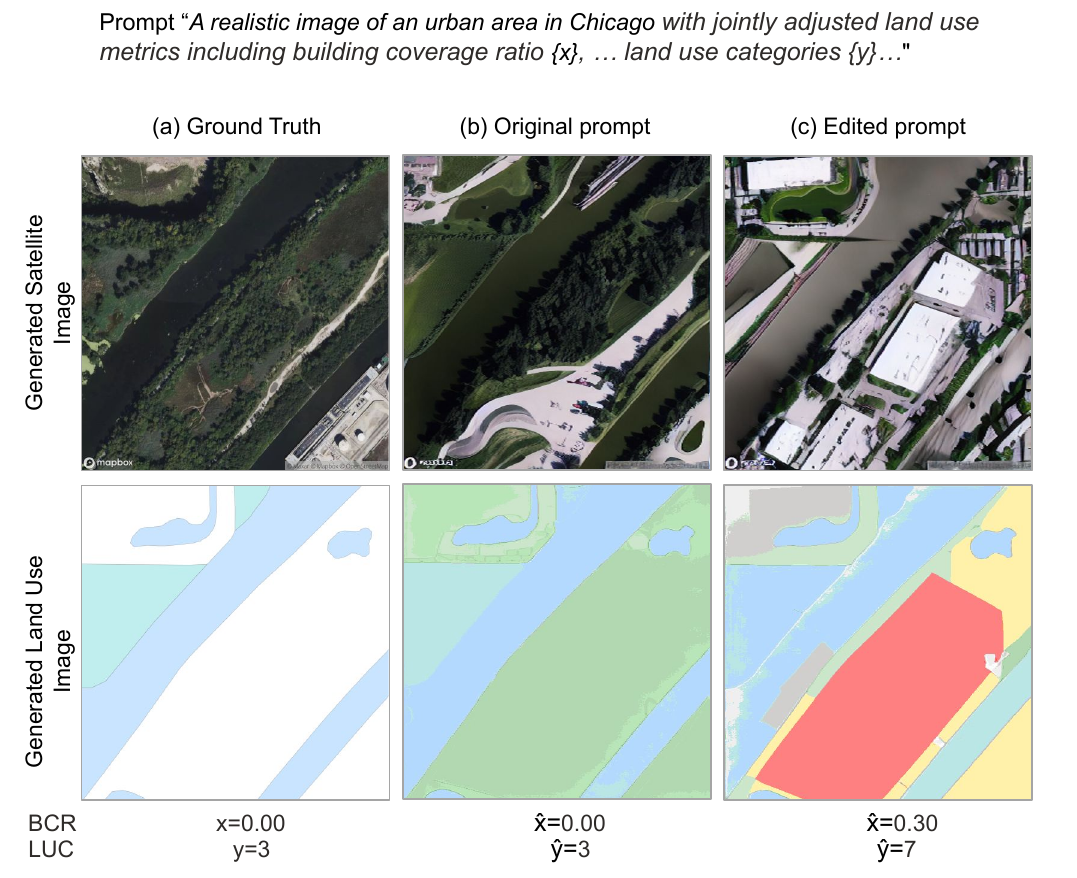}
    \caption{Generative scenario planning in Chicago under joint adjustment of multiple land-use metrics. (a) Ground truth. (b) Original prompt. (c) Edited Prompt. BCR = building coverage ratio; LUC = land use categories. Both are calculated from generated land use maps.}
    \label{fig:sus1}
\end{figure}
Figure~\ref{fig:sus1} presents the Chicago site. Figure~\ref{fig:sus1}a shows the ground truth, in which the satellite image contains a river corridor with adjacent open land and few built structures, and the corresponding land-use map records three land use categories with large contiguous areas. Figure~\ref{fig:sus1}b shows the scenario generated under the original prompt, which specifies a building coverage ratio of 0.00 and three land use categories. The generated outputs return a measured building coverage ratio of 0.00 and three land use categories, reproducing the specified inputs.

Figure~\ref{fig:sus1}c presents the sustainable development scenario under multidimensional land-use controls. The measured building coverage ratio rises to 0.30 and the number of land use categories increases from three to seven, confirming quantitative alignment between the planning objectives and the generated outputs. Spatially, a commercial zone emerges along the riverside and forms a dominant built mass that reorganizes the waterfront. Building footprints expand and convert previously open terrain into compact development parcels. Surrounding this commercial core, apartment-style residential neighborhoods appear in organized blocks supported by a more structured road layout that subdivides the site into accessible parcels. The gain in diversity is expressed through functional zones beyond the commercial core, while green spaces and water bodies remain preserved along the river edge, maintaining environmental continuity under denser development. This combination of intensified development and preserved environmental buffers is precisely the trade-off that planners negotiate in waterfront redevelopment, and the model renders it as a visible and measurable land-use map rather than an abstract target.

\begin{figure}[ht!]
    \centering
    \includegraphics[width=0.9\linewidth, keepaspectratio]{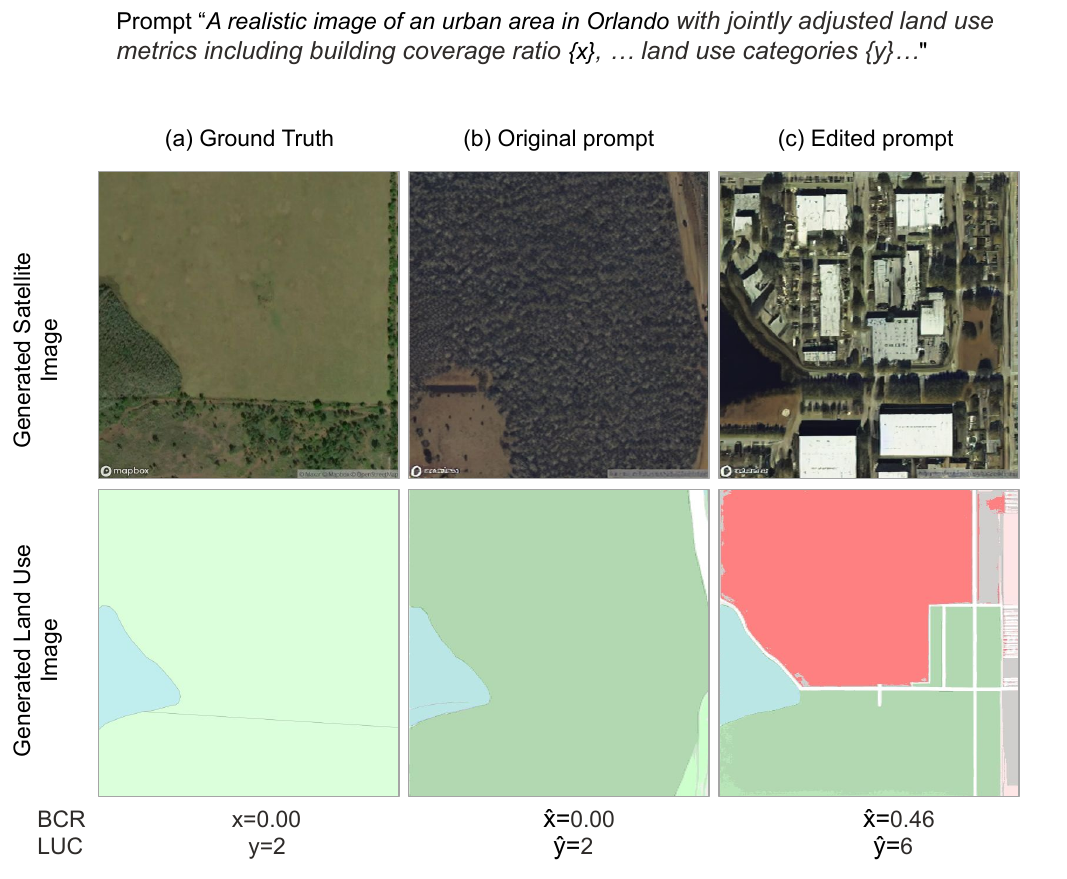}
    \caption{Generative scenario planning in Orlando under joint adjustment of multiple land-use metrics. (a) Ground truth. (b) Original prompt. (c) Edited Prompt. BCR = building coverage ratio; LUC = land use categories. Both are calculate from generated land use maps.}
    \label{fig:sus2}
\end{figure}

Figure~\ref{fig:sus2} presents the Orlando site. Figure~\ref{fig:sus2}a shows the ground truth, in which the satellite image contains grassland or meadow areas adjacent to water bodies with few built structures, and the land-use map records two land use categories. Figure~\ref{fig:sus2}b shows the scenario generated under the original prompt, which specifies a building coverage ratio of 0.00 and two land use categories, and the generated outputs return the same values. The generated satellite image renders the vegetated surface with a woodland texture rather than the lighter grassland appearance of the ground truth, and the category boundaries in the generated land-use map do not align exactly with the subtle texture transitions in the satellite image. This slight inconsistency suggests that the model may generalize vegetated areas toward dominant patterns in the training data, indicating a limitation in fine-grained generation. 

Figure~\ref{fig:sus2}c presents the sustainable development scenario under multidimensional land-use controls, in which the building coverage ratio prompt increases from 0.00 to 0.45 and the number of land use categories from two to five. The measured building coverage ratio in the generated land-use map reaches 0.46, closely matching the specified density target, while the measured number of land use categories reaches six, slightly exceeding the specified diversity target. The generated outputs organize the new development into two distinct zones. In the upper portion of the site, the satellite image shows a compact cluster of mid-sized flat-roofed buildings served by paved lots and access drives, and the land-use map represents this zone as a dominant commercial polygon that occupies most of the upper and central area. In the lower portion, the satellite image shows large-block warehouse structures with wide rectangular footprints set behind a road corridor, indicating a logistics or light-industrial typology. This land-use outcome is highly relevant to planning because it concentrates employment uses on previously undeveloped land while retaining an environmental buffer around the water, which is the spatial trade-off planners negotiate when siting commercial and industrial capacity. 

But we also observe some cross-modal misalignment in such multidimensional scenario generation. As shown in Figure~\ref{fig:sus2}c, the land-use map assigns the lower portion of the site to a green category, whereas the corresponding satellite image renders that same area as large-block warehouse facilities. The two modalities therefore disagree on the functional identity of the lower blocks. This misalignment suggests that the model might fail to propagate the built structures into the semantic categories of the land-use map, and it reinforces the need for the cross-modal validation that our land-use and spatial consistency metrics are designed to support.

Overall, our sustainable land-use development generation under multidimensional land-use controls demonstrates that DualControlNet can shape density, functional composition, and spatial configuration simultaneously. The generated outputs align with the prespecified numeric targets in both Chicago and Orlando, and they remain verifiable through the same land-use metrics in the prompts. This closes an important loop in the practice of scenario planning: a planner states multidimensional development goals, the model returns a spatially coherent and visually inspectable land-use map, and the map is then checked against the initial planning goals rather than general-purpose visual realism alone.

\section{Discussion}
\label{sec:dis}
Our study demonstrates DualControlNet as a generative land-use model, as opposed to the predominant predictive land-use models. Empirically, it synthesized land-use maps that do not exist in the training data and yet match the prescribed planning objectives for both Chicago and Orlando. Such a generative land-use model offers several capabilities that predictive land-use models cannot offer. First, it is controlled by textual prompts, so a planner can specify ``high-density development'' rather than a numeric transition matrix. Second, it directly returns visual outcomes, namely a land-use map paired with satellite-like imagery, which renders an abstract density target as an intuitive urban map. Third, it can generate diverse outcomes: under an identical density prompt, the model produced single-family, commercial, and industrial morphologies (Figure~\ref{fig:alt}), presenting planners with a portfolio of alternatives rather than a single predicted trajectory. Predictive land-use models, whether cellular automata, vector-based, or learning-based \citep{He2023, CruzBello2024, Guan2024, Yao2024, Cai2026}, forecast numeric transitions among existing land-use categories and cannot synthesize the final land-use maps and satellite imagery. Therefore, we complement the predictive models with a generative one, shifting the forecasting question to envisioning, rendering, and comparing multiple land-use scenarios.

Our study also demonstrates a control-verification cycle in the generation process. In DualControlNet, land-use metrics serve a dual role. First, they are control factors: embedded in textual prompts, they specify the planning objective that guides generation. Second, they are verification criteria: because the model returns a land-use map rather than satellite imagery alone, the same metrics can be recomputed from the generated output and compared against the initial objectives. The generated land-use maps recovered the prescribed density and diversity targets in both Chicago and Orlando, closing a cycle among planning objectives, generation, and post-generation verification. A satellite-only model cannot close this cycle, because imagery without semantic land-use categories yields no land-use metric for verification, however realistic it appears. We suggest that this cycle of controlled and verified generation should extend beyond land use. It applies to any domain whose objectives can be expressed as interpretable metrics that are both specifiable in a prompt and recomputable from the generated artifact. Examples can include connectivity and accessibility in transportation network design, floor area ratio in building design, or even habitat connectivity in ecological planning. In each case, the generated outcomes are verified against the domain's own criteria rather than generic metrics \citep{castro2025comprehensive, ye2022masterplangan, abdollahi2020building}. In short, we seek to make domain-specific metrics serve as both the control interface and the verification standard. A generative approach without domain-specific verification wouldn't be trustworthy for planning practice.

The multimodal design of DualControlNet carries further implications for planning practice. The two generated outcomes are complementary: land-use maps describe functional categories and provide semantic structure, while satellite imagery conveys texture, morphology, building patterns, and other visual characteristics. This multimodal generation matters because the same land-use category can contain substantial morphological heterogeneity. As shown in Figure~\ref{fig:alt}, under an identical building coverage prompt, the model can generate three typologies, all satisfying the same numeric target. A predictive model reporting the same coverage ratio would collapse these three futures into one number, let alone the subtle differences conveyed through satellite imagery. By jointly generating both layers, DualControlNet represents visual richness while retaining a semantic anchor for verification, extending beyond prior generative urban research \citep{ye_masterplangan_2022, quan_urban-gan_2022, jiang_generative_2024}. The geospatial consistency loss holds the two layers together. In fact, other consistency losses are also possible, and as urban systems always consist of multiple interdependent layers, future work may explore additional cross-modal constraints across transportation, building, land use, and environmental layers. 

Beyond technical aspects, the generative land-use models could lower barriers to public participation and planning communication. Predictive land-use models have long been used to support public engagement, but they are often difficult for non-experts to interpret. Such models require extensive data preparation, calibration, and parameter tuning that can exceed the institutional capacity of planning agencies \citep{vonk_bottlenecks_2005, vonk_improving_2008}. On the contrary, the prompt-based generation of land-use maps offers a lower-friction entry point: a stakeholder can state ``high-density development'' and receive a visual land-use map within seconds. This implication is consistent with prior work that argues for AI's potential to broaden participation by helping non-experts generate and compare design alternatives \citep{quan_urban-gan_2022, jiang_generative_2024}. Residents, policymakers, and community groups may not easily interpret land-use entropy, intersection density, or building coverage ratios, but they can readily discuss a rendered satellite image and its corresponding land-use map. Therefore, generative AI can thus support deliberation by translating abstract planning indicators into visual representations. 

Several limitations are worth acknowledging. First, the generated outcome pairs sometimes show cross-modal mismatches. In the multidimensional generation, the land-use map assigns the lower blocks to a green category while the satellite image renders the same area as warehouse facilities (Figure~\ref{fig:sus2}c). We also observe semantic confusion in transition areas, where wetland or water-adjacent spaces are represented as built-up or industrial land uses with irregular building blocks. These limitations actually echo land-use simulation research, which shows that parcel structure, within-parcel variation, and neighborhood relationships affect spatial validity \citep{Guan2024, Yao2024}. Second, our land-use labels are derived from OpenStreetMap, which is not sufficiently fine-grained in undeveloped areas. As a result, the input labels can be inconsistent across regions. Finer sources such as official plans, parcel maps, and points of interest would improve label fidelity. Third, the dual generation pipeline is not fully stable, occasionally producing map flips and orientation errors, and generated tiles are not continuous across tile boundaries. These limitations suggest that future work should explore additional cross-modal constraints, boundary-aware generation, and more robust training strategies to improve the quality of land-use map generations.

\section{Conclusions}
\label{sec:con}
Land-use patterns shape urban sustainability, health, and safety. However, existing land-use models are largely predictive, forecasting incremental transitions among established land-use categories from historical data, while lacking the capacity to envision land-use patterns that do not yet exist. Meanwhile, existing generative models remain unimodal, weakly controllable, and evaluated mainly by generic image metrics, being insufficient for land-use scenario planning. To address these gaps, we developed DualControlNet, a generative land-use model that jointly synthesizes land-use maps and satellite imagery, controls generation with interpretable land-use metrics such as building coverage ratio and land-use diversity, and verifies the generated outputs against the same metrics. We trained the model on more than 40,000 paired tiles from Orlando and Chicago and evaluated it along land-use, spatial consistency, and visual realism dimensions. This study yields the following findings.

First, joint multimodal generation substantially improves the consistency between the generated land-use maps and the targeted land-use metrics. Relative to the unimodal ControlNet baselines, DualControlNet raises the consistency in land-use diversity by 62.8\%, in land-use area by 49.5\%, in building coverage ratio by 16.3\%, and in distribution similarity by 5.9\%. Notably, the satellite-only baseline cannot be verified on any land-use metric, because imagery alone carries no semantic land-use categories. Second, the geospatial consistency loss strengthens cross-modal alignment. Both DualControlNet variants roughly double the cross-modal alignment of the unimodal baselines, and adding the loss further improves gradient-based cosine similarity and direction distribution similarity, while also lowering the Fréchet Inception Distance. Multimodal joint generation thus enhances not only cross-modal coherence but also perceptual image quality. With the technical achievements, we conducted scenario-based generative experiments demonstrating that DualControlNet responds to both single and multiple land-use controls. Under single-dimensional control, the BCR measured from the generated land-use maps tracks the prompted targets closely, while still producing residential, commercial, and industrial morphologies under an identical density prompt. Under multidimensional control, the model raises density and diversity together at underdeveloped sites in both Chicago and Orlando, generating riverside commercial cores and employment clusters that meet the specified targets while preserving environmental buffers. Overall, these findings show that land-use metrics can serve simultaneously as the control interface and the verification standard for generative land-use modeling. 

This study suggests that generative AI can complement predictive land-use modeling: it shifts the planning question from forecasting a single trajectory to envisioning, rendering, and comparing multiple land-use maps. It potentially lowers the technical barrier by allowing stakeholders to state an objective in plain language and inspect a visually explicit scenario within seconds. Several limitations remain, as discussed in Section~\ref{sec:dis}: residual cross-modal misalignment, coarse land-use labels in undeveloped areas, and discontinuity of the generated tiles at their boundaries. Future work should fold these failure modes into the control--verification loop by developing boundary-aware, topology-aware, and semantics-aware verification mechanisms, and by extending the framework to finer land-use sources and additional cities. Overall, this study presents a generative land-use modeling framework that is multimodal, controllable, and verifiable, and it demonstrates the potential of generative AI to support urban planning practice.

\section*{Data and Codes Availability Statement}
The anonymized data and Python scripts that support the findings of this study are available in a GitHub repository:{\url{https://github.com/qingqisqq/DualControlNet}}.

\bibliography{0_Paper_gen_ai_Qingqi.bib,TraditionalBEStudies.bib,GenAIandDiffusionModels.bib,trb_template,enrich_multiview,enrich_urbandif,land_use_modeling,enrich_scenario}
\newpage
\appendix
\section{Appendix A}

\begin{figure}[H]
    \centering
    \includegraphics[width=0.9\linewidth, keepaspectratio]{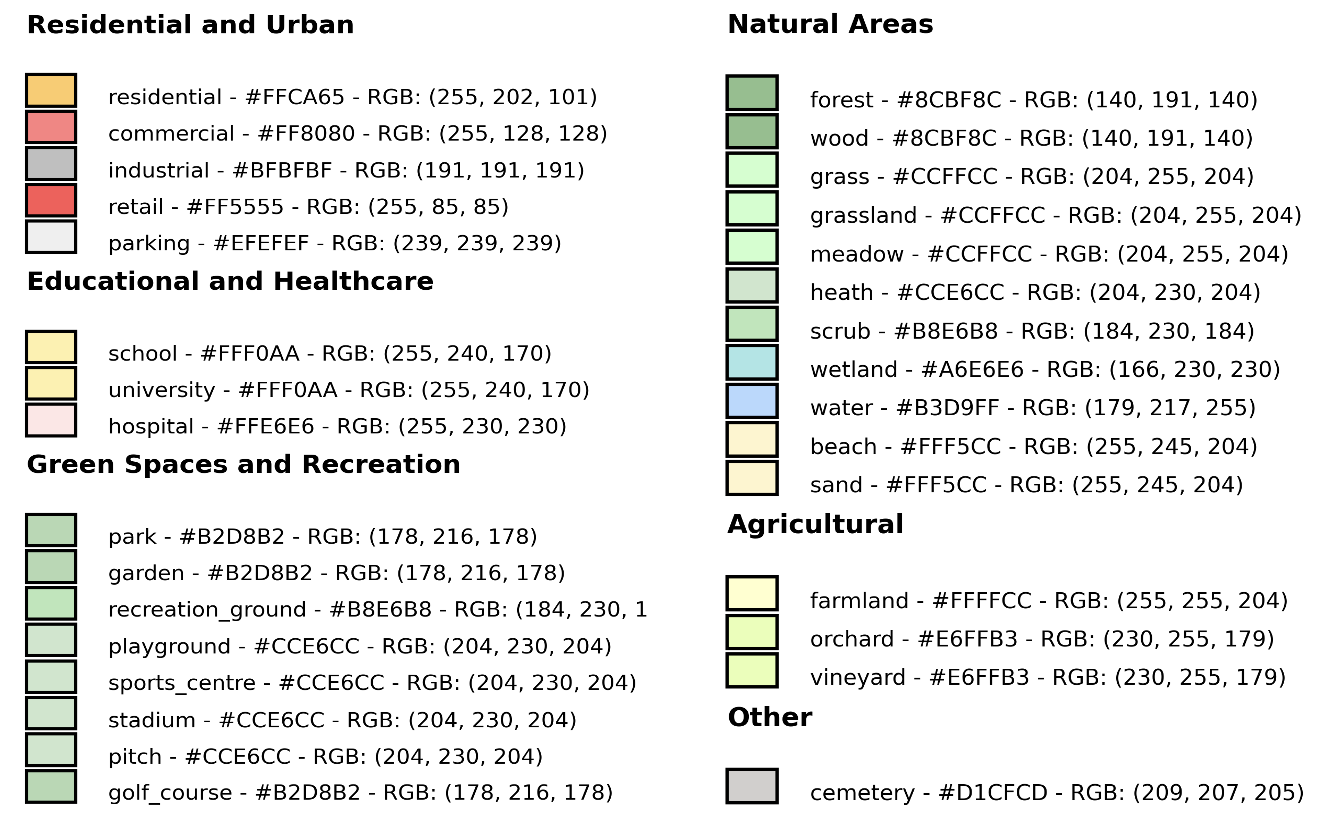}
    \caption{Land Use Legend Demonstration}
    \label{fig:legend}
\end{figure}
\end{document}